\documentclass[sigconf]{acmart}

\usepackage{booktabs} 

\acmConference[KDD'17]{ACM KDD conference, Data-Drive Discovery Workshop}{August 2017}{Halifax, NS, CANADA} 

\begin{document}
\title{Biomedical Document Clustering and Visualization based on the Concepts of Diseases}

\author{Setu Shah}
\affiliation{%
  \institution{Purdue School of Engineering and Technology, IUPUI}
  \streetaddress{799 W. Michigan Street}
  \city{Indianapolis} 
  \state{Indiana} 
  \postcode{46202}
}
\email{setshah@iupui.edu}

\author{Xiao Luo}
\affiliation{%
  \institution{Purdue School of Engineering and Technology, IUPUI}
  \streetaddress{799 W. Michigan Street}
  \city{Indianapolis} 
  \state{Indiana} 
  \postcode{46202}
}
\email{luo25@iupui.edu}


\begin{abstract}
Document clustering is a text mining technique used to provide better document search and browsing in digital libraries or online corpora. A lot of research has been done on biomedical document clustering that is based on using existing ontology.  But, associations and co-occurrences of the medical concepts are not well represented by using ontology. In this research, a vector representation of concepts of diseases and similarity measurement between concepts are proposed. They identify the closest concepts of diseases in the context of a corpus. Each document is represented by using the vector space model. A weight scheme is proposed to consider both local content and associations between concepts. A Self-Organizing Map is used as document clustering algorithm. The vector projection and visualization features of SOM enable visualization and analysis of the clusters distributions and relationships on the two dimensional space. The experimental results show that the proposed document clustering framework generates meaningful clusters and facilitate visualization of the clusters based on the concepts of diseases.

\end{abstract}

%
%



\maketitle

\section{Introduction}

Active research in the medical and biomedical domain has generated pervasive documents and articles. MEDLINE, the largest biomedical text database, has more than 16 million articles. It is estimated that more than 10,000 articles are added to MEDLINE weekly \cite{Yoo:2006}. There are continuously needs for development of techniques to discover, search, access and share knowledge from these documents and articles. Text clustering techniques enable us to group similar text documents in an unsupervised manner. 

Most of the research related to the biomedical document clustering focuses on either reforming the representation of biomedical documents or improving the clustering algorithms. Biomedical document clustering is different from the general text document clustering task, because in the latter, semantic similarities between words or phrases are not usually considered. One medical concept of disease might be represented in different forms, and some medical concepts of diseases might be highly correlated. For example, `Type 2 Diabetes' is the same concept of disease as `Diabetes Mellitus Type 2'.  `Hypertension' might co-occur often with `Stroke'. In order to capture the semantic similarities between words or phrases, previous research on document representation reforming  \cite{Logeswari:2013} \cite{Yoo:20061} \cite{Zhang:2007} often use existing ontology such as MeSH or WordNet to identify the semantic relationships. However, ontology doesn't reflect the co-occurrences of medical concepts. This paper focuses on biomedical document clustering based on the concepts of diseases. The proposed similarity measure between the concepts of diseases is based on the Word2vec model \cite{Mikolov:2013}. This similarity measure identifies the closest concepts based on co-occurrences of the concepts. The proposed concept weighting scheme is the linear combination of the TF-IDF value which reflects the content similarity between documents and the similarity score based on the proposed similarity measurements that reflect the semantic similarity between documents. 

The unsupervised learning algorithm Self-Organizing Map (SOM) \cite{Kohonen:1998} has been used as the clustering technique. SOM has properties of both vector quantization and vector projection. The neurons of an SOM can be presented on a two dimensional space. By projecting the input data instances to their best matching units (BMUs) on the SOM map, the distribution of the inputs can be visualized on the two dimensional space with the U-matrix and hit histogram of the SOM. The relationships of the clusters based on the concepts of diseases can be visualized. This clustering visualization is a beneficial feature for biomedical literature search and browsing based on concepts of diseases. 

The rest of the paper is organized as followings. In section 2, related work is described. Section 3 demonstrates how the concepts of diseases are extracted by using UMLS MetaMap. Section 4 and 5 detail the measurement of concepts similarity and weighting scheme for each concept in the document representation. Experimental settings and results are given in section 6. Section 7 concludes his research and discusses potential future work.

\section{Related Work}
A lot of research has been done in biomedical document clustering in past decades. Some of it focused on document presentation reforming based on medical ontology or on using different weighting scheme other than TF-IDF, while some others focused on investigating various clustering algorithms.

Zhang et al. \cite{Zhang:2007} reviewed three different ontology based term similarity measurements: path based \cite{Palmer:1994}, information content based \cite{Resnik:1999}, and feature based \cite{Knappe:2007} and then proposed their own similarity measurement and term re-weighting scheme. K-means algorithm is used for document clustering. Based on the results comparison, some of them are slightly worse than the word based scheme. The authors mentioned that it might because of the limitation of the domain ontology, term extraction and sense disambiguation. Visualization of the relationships between the clusters were not included in this research.

Yoo et al. \cite{Yoo:2006} used a graphical representation method to represent a set documents based on the MeSH ontology, and proposed the document clustering and summarization with this graphical representation. The document clustering and summarization model gained comparable results on clustering and also provided some visualization on the documents cluster model based on the relationships of the terms. However, this visualization relies largely on the MeSH ontology instead of the document relationships themselves.

Logeswari et al. \cite{Logeswari:2013} proposed a concept weighting scheme based on the MeSH ontology and tri-gram extraction to extract concepts from the text corpus. The semantic relationship between tri-grams are weighted through a heuristic weight assignment of four predefined semantic relationships.  The K-means clustering algorithm results show that concept based representation was better than word based representation. Visualization of the clustering results was not investigated.

Gu et al. \cite{Gu:2013} proposed a concept similarity measurement by using a linear combination of multiple similarity measurements based on MeSH ontology and local content which include TF-IDF weighting and co-efficient calculation between related article sets. A semi-supervised clustering algorithm was employed at the stage of document clustering. Their focus was not clustering visualization.

Some research has been done about the visualization process to support biomedical literature search. Gorg et al. \cite{Gorg:2010} developed a visual analytics system, named Bio-Jigsaw by using the MeSH ontology. This research demonstrated how visual analytics can be used to analyze a search query on a gene related to breast cancer. Neither document representation nor document clustering were discussed.

To the best of authors' knowledge, this research is the first to present concepts of diseases using vectors based on the Word2vec model instead of using an ontology. The proposed similarity measurement and the concept weighting scheme are first applied to the biomedical document clustering. The SOM based clustering is employed to visualize the distribution of document clusters based on the concepts of diseases.

\section{Concepts of Diseases Extraction}
In this work, the focus is on clustering biomedical documents based on the concepts of diseases that are addressed by or mentioned in the documents. To extract the concepts of diseases from the documents, Unified Medical Language System (UMLS) MetaMap is used. UMLS MetaMap  \cite{MetaMap} is a natural language processing tool that makes use of various sources such as UMLS Metathesaurus \cite{Metasaus} and SNOMED CT \cite{SCT} to map the phrases or terms in the text to different semantic types. 

\begin{figure}
\centering
\includegraphics[scale=0.45]{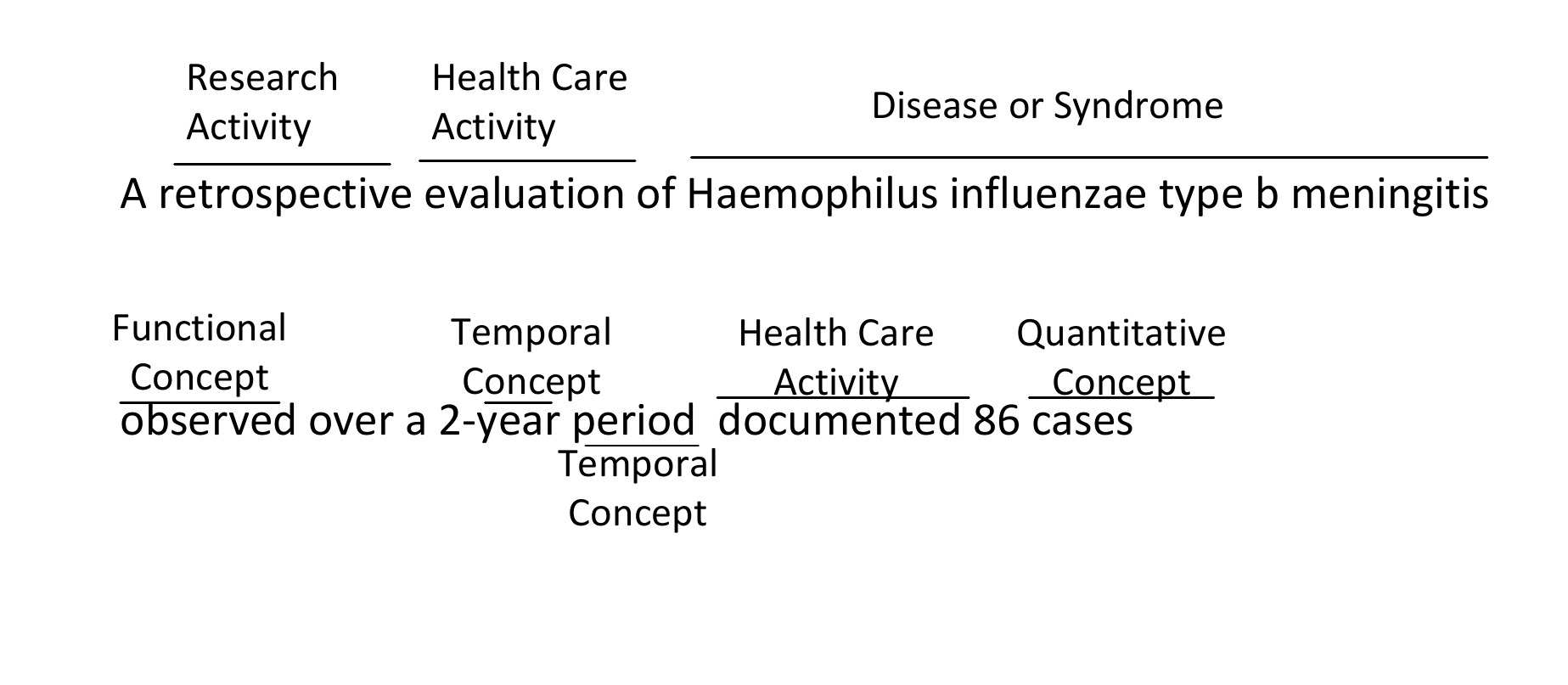}
\caption{Semantic Mapping of using UMLS MetaMap}
\label{MetaMap1}
\end{figure}

Figure \ref{MetaMap1} provides an example of mapping phrases to different semantic types using UMLS MetaMap. In this example, eight terms or phrases in the sentence have been mapped to six semantic types. The phrase `Haemophilus influenzae type b meningitis' in the sentence has been identified as semantic type `disease or syndrome' and mapped to phrase `Type B Hemophilus influenzae Meningitis' based on the lexicon that UMLS MetaMap uses. In this research, if a term or phrase has been mapped to semantic types `Disease or Syndrome' or `Neoplastic Process', the corresponding phrase in the lexicon produced by MetaMap is extracted.

\section{Concepts Similarity Measure} \label{s3}
In the biomedical literature, same concepts of a disease can be presented by different terms or combinations of words. For example, `cancer of breast' and ` breast cancer' are two phrases that present the concept of the same disease. However, they are treated as two different concepts if typical vector space model and TF-IDF weighting scheme are used for document presentation, and the semantic similarity between them is not measured. In this research, a semantic similarity measure between different concepts of diseases is proposed. Given a total of $L$ concepts of diseases extracted from the raw text corpus, the similarities between any two concepts are stored in the similarity matrix $S$ as presented in Equation \ref{smatrix}. Each entry $s_{i,j}$ in the matrix $S$ represents the similarity between concept $C_i$ and $C_j$.

\begin{equation}
S_{L,L} = 
 \begin{pmatrix}
  s_{1,1} & s_{1,2} & \cdots & s_{1,L} \\
  s_{2,1} & s_{2,2} & \cdots & s_{2,L} \\
  \vdots  & \vdots  & \ddots & \vdots  \\
  s_{L,1} & s_{L,2} & \cdots & s_{L,L} 
 \end{pmatrix}
\label{smatrix}
\end{equation}
 
To calculate the similarity between two concepts, first, each word is represented by a vector (as proposed in Equation \ref{WVec}). This vector representation is learned by training the Word2Vec model. The Word2Vec training algorithm was developed by a team of researchers at Google led by Tomas Mikolov \cite{Mikolov:2013}. It is a computationally-efficient algorithm to generate vectors of real numbers to present words in a given raw text corpus. These vector representations are learned using three-layer neural networks using either a continuous bag-of-words approach or a skip-gram architecture. The vectors preserve the distances between words in the vector space so that the words that share common contexts in the raw text corpus are located in close proximity to one another. The dimension of the vector created depends on the number of neurons in the hidden layer of the neural network when training a Word2Vec model.

\begin{equation}
Word = (wv_1, wv_2, \dots, wv_m)
\label{WVec}
\end{equation}
$m$: the dimension of the vector.

In this research, a trained Word2Vec model \cite{Moen:2013} created from a subset of PubMed literature database and a subset of PubMed Central (PMC) Open Access database is employed. These two text corpus contains a large number of biomedical documents. The trained model creates 200 dimensional vectors to present the words extracted in the two text corpus. The skip-gram architecture with a window size of 5 is adopted for the learning process \cite{Moen:2013}.

Although some of concepts of diseases contain only one word, many of them span multiple words. In this work, if a concept of disease spans multiple words, a concept vector is generated by aggregating the vectors of all the words in the concept, as shown in Equation \ref{phrase-vectors}. For example, for the disease `diabetes mellitus', the vector for `diabetes' and the vector for `mellitus' are aggregated by adding them together.

\begin{equation}
C = \sum_{i=1}^M Word_i
\label{phrase-vectors}
\end{equation}
 
$M$: the total number of words in a concept $C$.
 
The similarity score between the concepts are calculated using the cosine distance between the vectors as shown in Equation \ref{cos-sim}.

\begin{equation}
\label{cos-sim}
S_{i,j} = \frac{\mathbf{C_i} \cdot \mathbf{C_j} }{\sqrt{\sum_{i=1}^{m} C_i^2}\cdot \sqrt{\sum_{i=1}^{m} C_j^2}} 
\end{equation}

By presenting concepts in vector and using this similarity measure, it is observed that the more the diseases are  associated, the higher the similarity scores between them are. Table \ref{phrase-vector-distances} provides some examples of concepts of diseases and the their top 3 closest concepts based on the similarity scores. `Hypertension' is often associated with `hyperlipidaemia' in the literature, so the similarity between them is higher than that between `Hypertension' and other concepts of diseases.

\begin{table}
\setlength{\tabcolsep}{1.5pt}

\caption{Examples of concepts and the top 3 closest concepts based on the similarity scores }
\label{phrase-vector-distances}
\begin{tabular}{| l | l | l |}
    \hline
	\textbf{Concept} & \textbf{Closest Concepts} & \textbf{Score of} \\
	& & \textbf{Similarity} \\
	\hline
		& essential hypertension & 0.813 \\
	hypertension & hyperlipidaemia & 0.692 \\
		& dyslipidemia & 0.659 \\
	\hline
	endothelial & dysfunction & 0.739 \\
	 dysfunction & renal dysfunction & 0.660 \\
		& cortical dysfunction & 0.639 \\
	\hline
	carpal tunnel	& bilateral carpal tunnel syndrome & 0.970 \\
	syndrome & cts carpal tunnel syndrome & 0.957 \\
		& carpal tunnel & 0.941 \\
	\hline
		& diabetes mellitus & 0.918\\
	diabetes & diabetes mellitus type ii & 0.868\\
		& dm diabetes mellitus & 0.845\\
	\hline
	cardiovascular & cardiac diseases & 0.8181 \\
	disease & metabolic diseases & 0.8179 \\
		& heart diseases & 0.787 \\
	\hline	
\end{tabular}
\end{table}

\section{Document Representation and Weighting Scheme}
In this research, the typical vector space model is used to present a biomedical document, each entry of the vector corresponding to a concept of disease which is identified through the UMLS MetaMap. The proposed weight ($Weight_{C_{i,d}}$) that is given to each concept ($C_{i,d}$) is calculated as equation \ref{ws}:

	\begin{equation}
	 Weight_{C_{i,d}} =
  \begin{cases}
    tf_{{C_i},d}\times\log\frac{|D|}{df_{{C_i}}} + \sum_{j=1}^M S_{i,j}      & \quad \text{if } tf_{{C_i},d} > 0 \\
    \sum_{j=1}^N \frac{N-(j-1)}{N} S_{i,j}  & \quad \text{if } tf_{{C_i},d} = 0\\
  \end{cases}
   \label{ws}
	\end{equation}

$df_{C_{i}}$: the number of documents in which concept ${C_i}$ occurs at least once

$tf_{C_{i,d}}$: frequency of concept ${C_i}$ in document $d$

$|D|$: total number of documents in the corpus

$S_{i,j}$: the similarity between ${C_{i}}$ and concept $C_{j}$ that both occur in document $d$. $C_{j}$ is the $j^{th}$ frequent concept in the document $d$.

$M$: the total number of concepts in document $d$. 

$N$: top $N$ closest concepts of ${C_i}$. In this research, $N = 3$.

If a concept occurs in a document, the weighting scheme uses the TF-IDF value to underline the occurrence of the concept in the local content. The $\sum_{j=1}^M S_{i,j}$ calculates the sum of similarity scores between the occurred concept $C_{i,d}$ and other concepts ($C_{j,d}$, $j=1,\dot,M$) that also occurs within the document. If a concept does not occur in the document, the weight is calculated by a weighted sum of the top 3 closest concepts ($C_{j,d}$, $j=1,\dot,3$) that appear in the document based on the similarities scores. By using this weighting scheme, the representation measures the occurrences of different representations of the same or similar concepts. For example, `diabetes' occurs in one document, but `diabetes mellitus' occurs in another document. By using the traditional TF-IDF weighting scheme, their values would be 0 for documents in which the concept does not appear. However, by using the proposed weighting scheme, they are weighted based on the similarity between the concept and its closest concepts. Thus, for the document that does not contain the concept `diabetes mellitus', instead of using 0, the similarity score between `diabetes mellitus' and other concepts that appear in the document is used.

\section{Clustering Algorithm}
Self-Organizing Map (SOM) is used for document clustering visualization \cite{kohonen2000self}. SOM implements the topologically ordered display of the data to facilitate understanding the structures of the input data set. It is also readily explainable and easy to visualize. The visualization of the multidimensional data is one of the main application areas of SOM \cite{Kohonen:1998}. These features make SOM an appropriate choice as a clustering algorithm for this paper.

A basic SOM consists of $M$ neurons located on a low dimensional grid (usually 1 or 2 dimensional) \cite{Kohonen:1998}. The algorithm responsible for the formation of the SOM involves three basic steps after initialization: sampling, similarity matching, and updating. These three steps are repeated until
formation of the feature map has completed. Each neuron $i$ has a $d$-dimensional prototype weight vector $W_{i}={W_{i1},W_{i1},...,W_{id}}$. Given $X$ is a $d$-dimensional sample data(input vector), the algorithm is summarized as follows:

\begin{itemize}

\item Initialization:

Choose random values to initialize all the neuron weight vectors $W_{i}(0), i=1,2,...,M,$ where $M$ is the total number of neurons in the map.

\item Sampling:

Draw a sample data $X$ from the input space with a uniform probability.

\item Similarity Matching:

Find the best matching unit (BMU) or winner neuron of $X$, denoted here by $b$ which is the closest neuron (map unit) to $X$ in the criterion of minimum Euclidean distance, at time step $n$ ($n^{th}$ training iteration).

	\begin{equation}
	b=\arg\min_{i}{||X -W_{i}(n)||}, i=1,2,...,M \label{BMUeq}
	\end{equation}
	
\item Updating:

Adjust the weight vectors of all neurons by using the update formula ~\ref{UpdateSOM}, so that the best matching unit (BMU) and its topological neighbors are moved closer to the input vector $X$ in the input space.

	\begin{equation}
	W_{i}(n+1)=W_{i}(n)+\eta(n)h_{b,i}(n)(X-W_{i}(n))
	\label{UpdateSOM}
	\end{equation}

Where $\eta(n)$ denotes the learning rate and $h_{b,i}(n)$ is the suitable neighborhood kernel function centered on the winner neuron.

The distance kernel function can be, for example, Gaussian:

	\begin{equation}
	h_{b,i}(n)=e^{-\frac{||r_{b}-r_{i}||^{2}}{2\sigma^{2}(n)}}\label{Gaussian}
	\end{equation}

Where $r_{b}$ and $r_{i}$ denote the positions of neuron $b$ and $i$ on the SOM grid and $\sigma(n)$ is the width of the kernel or neighborhood radius at step $n$. $\sigma(n)$ decreases monotonically along the steps as well. The initial value of neighborhood radius $\sigma(0)$ should be fairly wide to avoid the ordering direction of neurons to change discontinuously. $\sigma(0)$ can be properly set to be equal to or greater than half the diameter of the map. Formula~\ref{sigma0} gives the initial value of the neighborhood radius for a map of size $a$ by $b$.

	\begin{equation}
	\sigma(0)=\frac{\sqrt{a^{2}+b^{2}}}{2}\label{sigma0}
	\end{equation}

\item Continuation:

Continue with sampling until no noticeable changes in the feature map are observed or the pre-defined maximum number of iterations is reached.

\end{itemize}
The most commonly used visualization techniques of SOM are the U-Matrix and Hit histogram.

\begin{figure}[thpb]
  \centering
\includegraphics[height=1.8in]{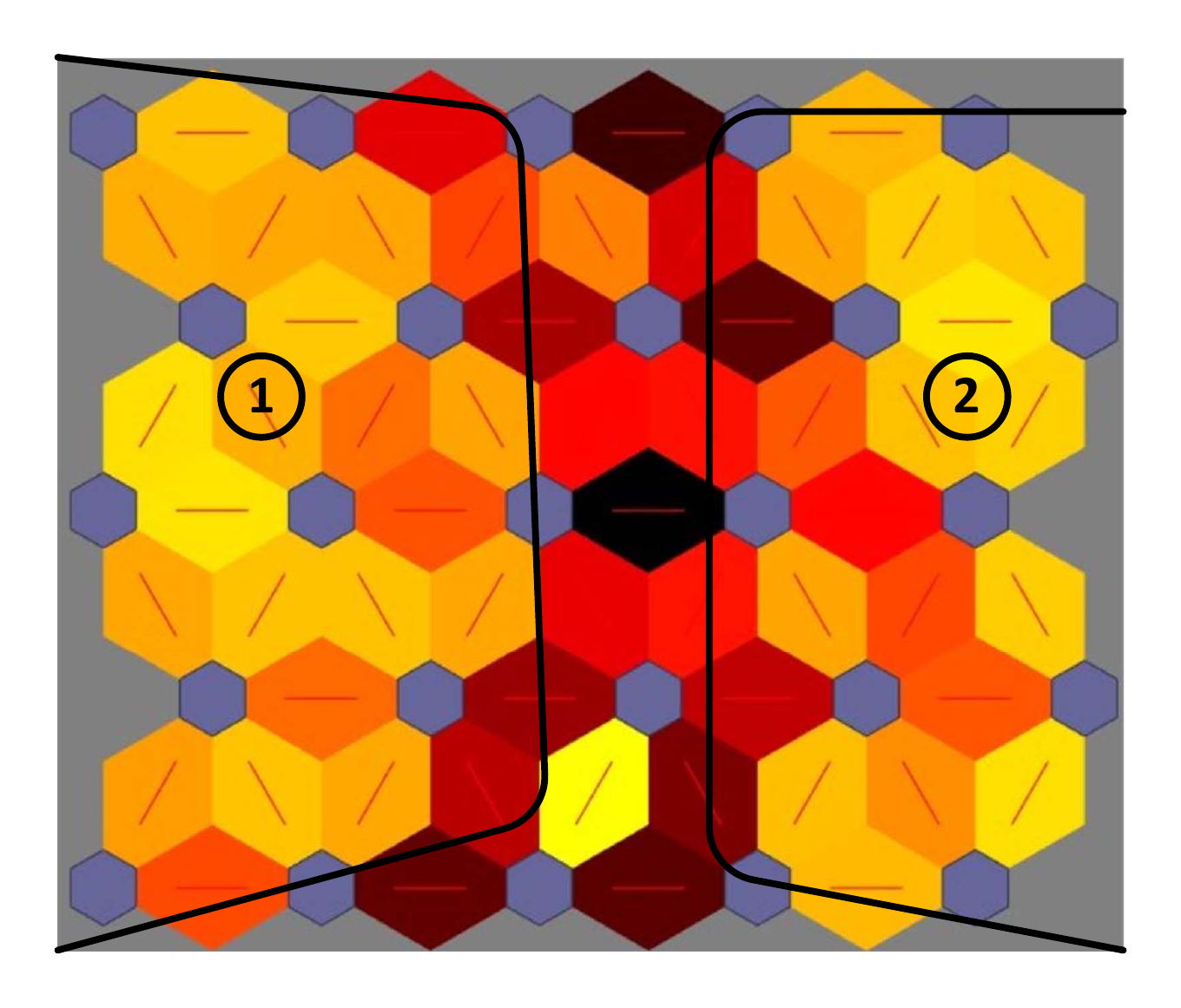}
  \caption{U-matrix of a trained SOM}\centering
  \label{UMatrixSOM}
\end{figure}

The U-matrix \cite{Kohonen:1998} holds all distances between neurons and their immediate neighbor neurons. Figure \ref{UMatrixSOM} shows the U-matrix of a trained map on a input data set that has two clusters. The lighter the color in the hexagon connecting any two neurons, the smaller is the distance between them. From the U-matrix, two large light regions can be visualized. One is towards the left, while the other is to the right. These regions present the two clusters obtained on training the input data set. The U-matrix gives a direct visualization of the number of clusters and their distribution.

\begin{figure}[thpb]
  \centering
\includegraphics[height=2.0in]{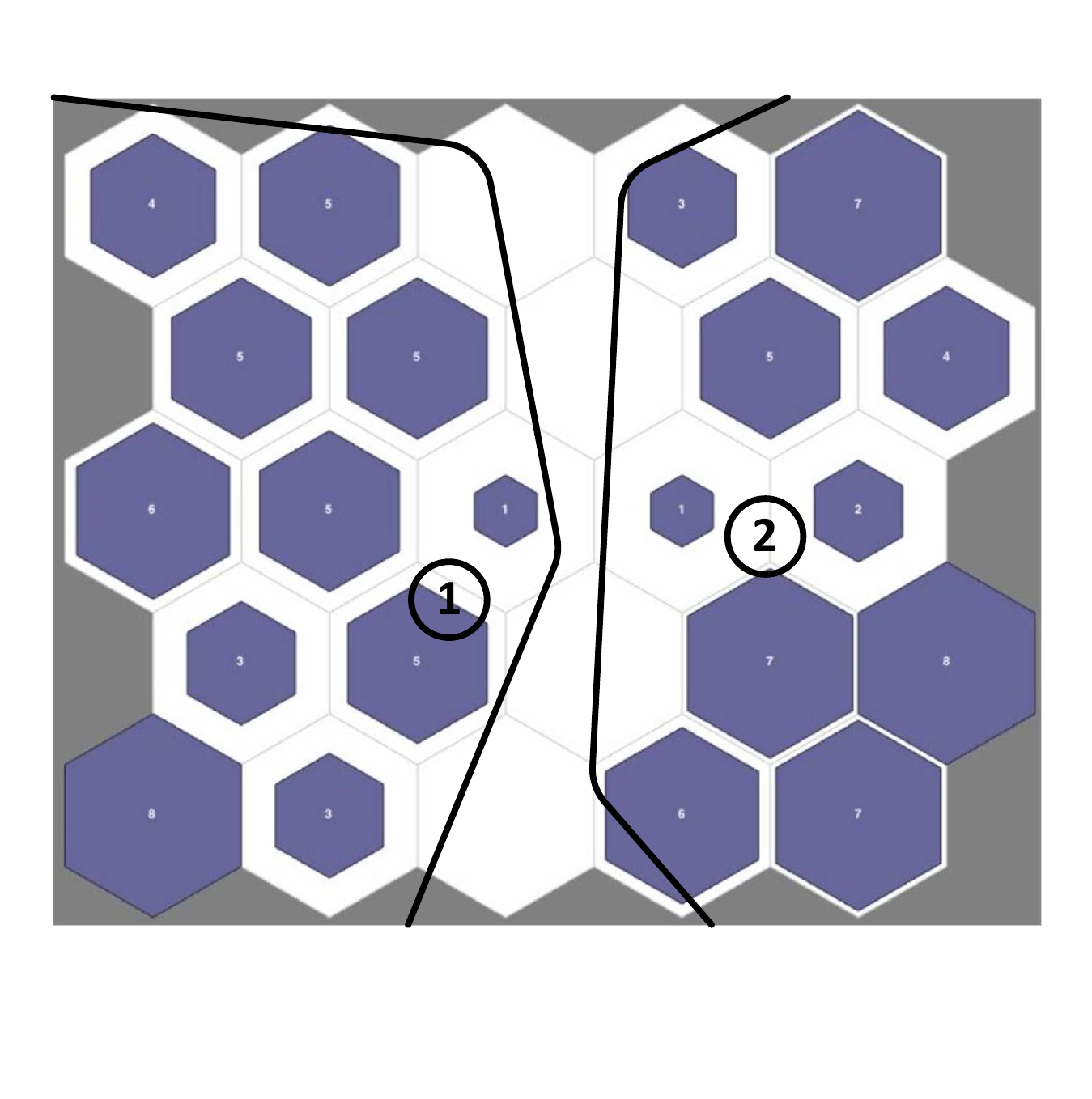}
  \caption{Hit histogram of a trained SOM}
  \centering\label{HitHistgramSOM}
\end{figure}

The hit histogram of the input data set on the trained map provides a visualization that details the distribution of input data across the clusters. Each input data instance in the data set can be projected to the closest neuron on a trained SOM map. The closest neuron is called the best matching unit (BMU) of the input data instance. The hit histogram is constructed by counting the number of hits each neuron receives from the input data set. Figure \ref{HitHistgramSOM} shows the hit histogram of an input data set on the trained SOM map. Each hexagon represents one neuron on the map. The size of the marker indicates the number of hits the neuron receives. Thus, a larger marker is representative of a larger number of hits on that neuron. Based on the hit histogram, it is visualized that most of the input data hits neurons in the left and right regions. These two regions correspond to the two clusters on the U-matrix shown in Figure \ref{UMatrixSOM}.


\section{Experiment Setting and Result Analysis}
To evaluate the proposed biomedical document clustering framework that is based on the concepts of diseases, two subsets of large biomedical document collections have been used: PubMed Central Open Access and Ohsumed collection. The details of these two document collections and the corresponding clustering results and visualization are detailed in the following subsections.
 
\subsection{Datasets}
PubMed Central Open Access data set has been used by many research projects to examine tasks of biomedical literature clustering and classification \cite{Zhang:2007} \cite{zhu2009enhancing}. It is also a part of the training corpus for the Word2Vec model used in this research. Ohsumed document collection is a data set that has been used by many researchers \cite{bloehdorn2006learning} \cite{simeon2008categorical} for text mining. Although the Ohsumed collection includes documents that are not up to date, it is used to evaluate the robustness of the proposed document clustering framework on a data set where concepts might be presented differently than the ones included in the training corpus for Word2Vec.

\subsubsection{\textbf{PubMed Central Open Access (PMC-OA)}}

The PubMed Central Open Access \cite{PMC-OA} is a subset of over 1 million articles from the total collection of articles in PMC. For this research, a set of 600 articles were randomly selected from the `A-B' subset which includes articles from journals whose names start with letter `A' or `B'. The number of selected articles from each journal is shown in Table \ref{journal-list}.

\begin{table}
\setlength{\tabcolsep}{1.5pt}
\caption{PMC-OA Data Set \label{journal-list}}
\begin{tabular}{| l | l |}
\hline
	\textbf{Name of journal} & \textbf{\# of documents} \\
	\hline
	American Journal of Hypertension & 13 \\
	\hline
	Augmentative and alternative communication & 2 \\
	\hline	
	Ancient Science of Life & 3 \\
	\hline
	Bioinformatics and biology insights & 45 \\
	\hline
	Allergy and asthma proceedings & 28 \\
	\hline
	BoneKEy reports & 4 \\
	\hline
	Anesthesia, essays and researches & 135 \\
	\hline
	Biological trace element research & 31 \\
	\hline
	Bone Marrow Research & 1 \\
	\hline
	Brain and language & 1 \\
	\hline
	American journal of physiology. & 11 \\
	Endocrinology and metabolism &  \\
	\hline
	Aphasiology & 3 \\
	\hline
	Annals of rehabilitation medicine & 323 \\
	\hline
\end{tabular}
\end{table}

To be consistent with the data set - Ohsumed Collection, only content in `Title' and `Abstract' sections from these documents are used. 658 unique concepts of diseases are identified by UMLS MetaMap. Figure \ref{words-in-concepts} shows the distribution of these concepts based on the number of words in each concept.

\subsubsection{\textbf{Ohsumed Collection}}

The Ohsumed collection \cite{ohsumed-collection} used here includes the abstracts of 20,000 articles. These articles are related to cardiovascular diseases and are further categorized into 23 cardiovascular disease categories. For this research, a subset of 600 documents is randomly selected. These documents cover all the 23 categories. Table \ref{ohsumed-categories} shows the number of documents selected from each category.

\begin{table}
\setlength{\tabcolsep}{1.5pt}
\caption{Ohsumed Collection \label{ohsumed-categories} }
\begin{tabular}{| l | l | l |}
\hline
	\textbf{Category} & \textbf{Label} & \textbf{\# of documents} \\
	\hline
	Bacterial Infections and Mycoses & C01 & 22 \\
	\hline
	Virus Diseases & C02 & 23 \\
	\hline
	Parasitic Diseases & C03 & 29 \\
	\hline
	Neoplasms & C04 & 26 \\
	\hline
	Musculoskeletal Diseases & C05 & 25 \\
	\hline
	Digestive System Diseases & C06 & 23 \\
	\hline
	Stomatognathic Diseases & C07 & 23 \\
	\hline
	Respiratory Tract Diseases & C08 & 24 \\
	\hline
	Otorhinolaryngologic Diseases & C09 & 29 \\
	\hline
	Nervous System Diseases & C10 & 23 \\
	\hline
	Eye Diseases & C11 & 28 \\
	\hline
	Urologic and Male Genital Diseases & C12 & 26 \\
	\hline
	Female Genital Diseases and & C13 & 27 \\
	Pregnancy Complications & & \\
	\hline
	Cardiovascular Diseases & C14 & 27 \\
	\hline
	Hemic and Lymphatic Diseases & C15 & 28 \\
	\hline
	Neonatal Diseases and Abnormalities & C16 & 25 \\
	\hline
	Skin and Connective Tissue Diseases & C17 & 28 \\
	\hline
	Nutritional and Metabolic Diseases & C18 & 27 \\
	\hline
	Endocrine Diseases & C19 & 30 \\
	\hline
	Immunologic Diseases & C20 & 26 \\
	\hline
	Disorders of Environmental Origin & C21 & 26 \\
	\hline
	Animal Diseases & C22 & 25 \\
	\hline
	Pathological Conditions, Signs &C23 & 29\\ 
	and Symptoms & & \\
	\hline
\end{tabular}
\end{table}

\begin{figure}
\centering
\includegraphics[scale=0.5]{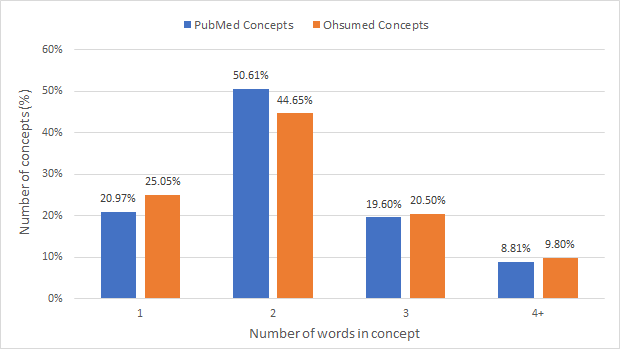}
\caption{Distribution of the concepts based on the number of words}
\label{words-in-concepts}
\end{figure}

After concepts of diseases are identified by using the UMLS MetaMap, 67 documents have no disease-related concept identified. These documents are not included in the experiments. In total, 1449 concepts of diseases are identified and extracted from the 533 documents. Figure \ref{words-in-concepts} shows the distribution of these concepts based on the number of words in the concepts in comparison with the concepts extracted from PMC-OA. Although the total number of concepts of diseases extracted from the Ohsumed collection is higher than that are extracted from the PMC-OA. The distribution based on the number of words in the concepts is very similar. Ohsumed collection has a slightly higher radio of concepts of one word whereas PMC-OA collection has a higher radio of concepts spanning two words. The percentages of concepts with 3 or more words are almost the same.

\subsection{Clustering, Visualization and Discussion}
SOM has been used for document clustering after concepts extraction and document representation using the proposed weighting scheme. The size of the map is 10 by 10 which contains 100 neurons. The training iterations are set to be 50,000. 

\begin{figure*}
\centering
\includegraphics[height=4.1in]{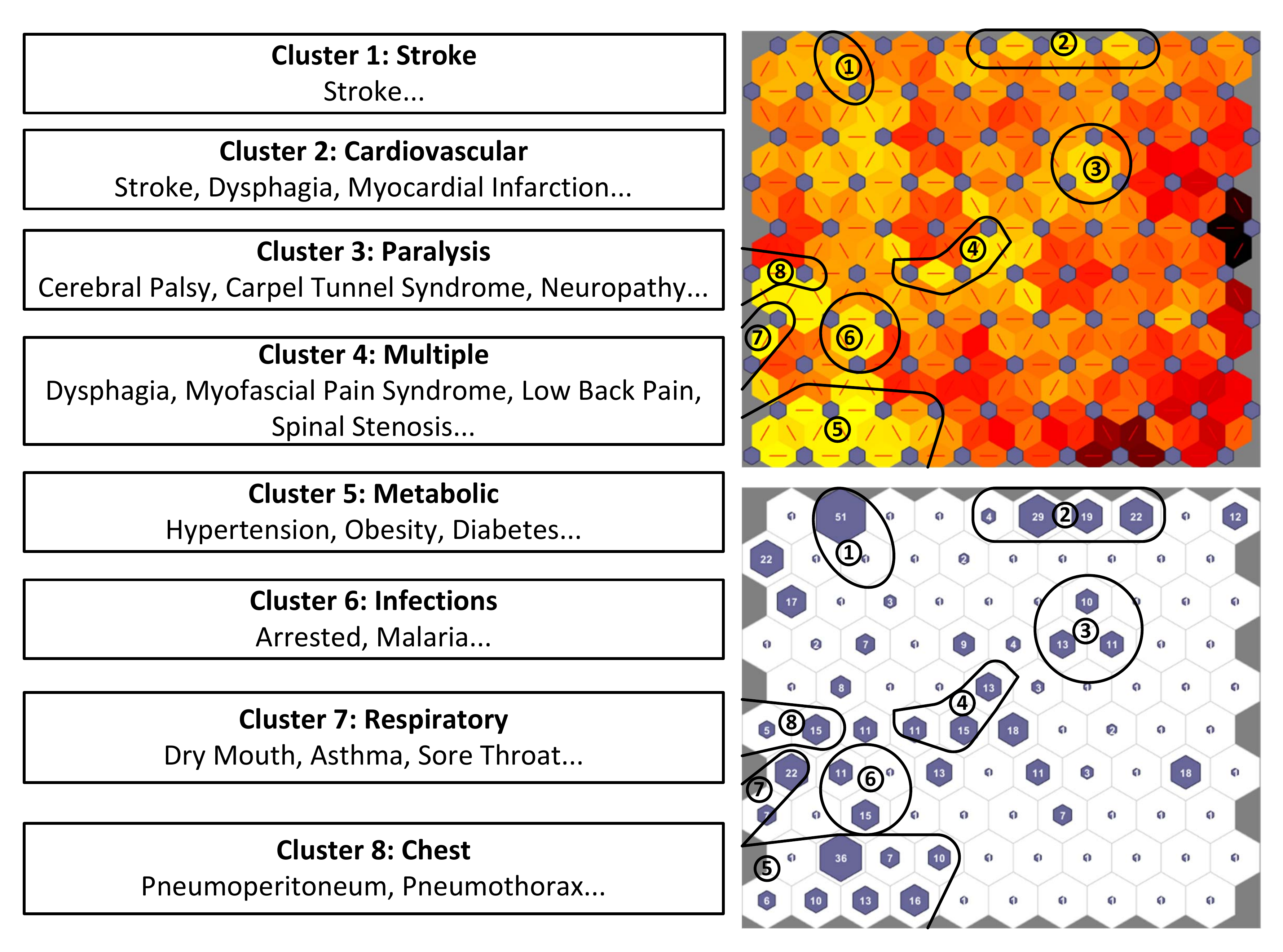}
\caption{Clustering results of PubMed Central Open Access}
\label{pubmeds}
\end{figure*}

\begin{figure*}
\includegraphics[height=3.75in]{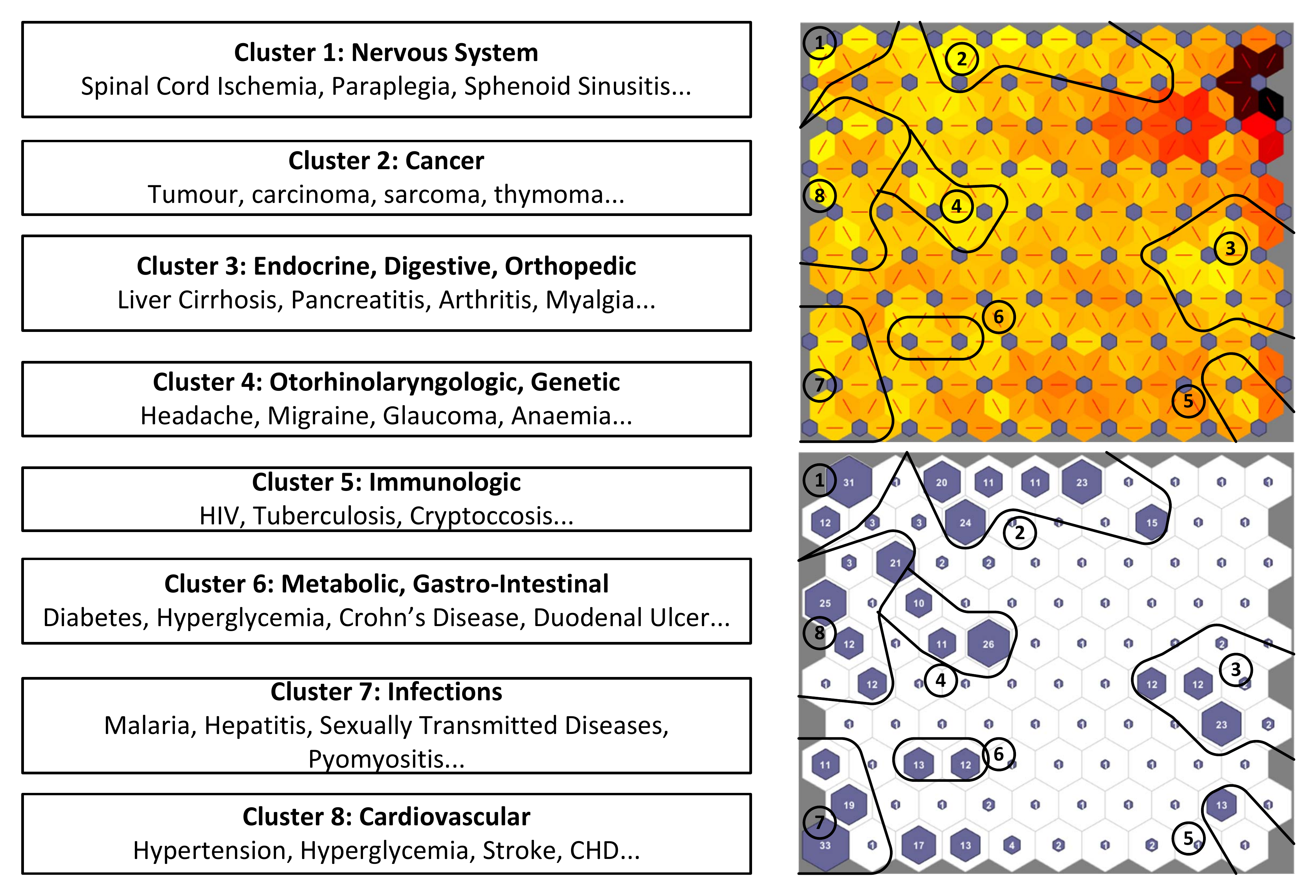}
\caption{Clustering results of Ohsumed Collection}\label{OHSU}
\end{figure*}

Figure \ref{pubmeds} shows the clustering results of document collection PMC-OA. Among 658 concepts that are extracted by using the UMLS MetaMap, 180 concepts occur in more than 1 document. Compare to the Ohsumed collection, a larger number of concepts have document frequency more than 1. That means the TF-IDF value in the weighting scheme has more impact on PMC-OA dataset than it does on the Ohsumed collection. The U-matrix of the trained SOM map shows more clear boundaries than that of the Ohsumed collection.

From the U-matrix and hit histogram, 8 clusters can be clearly identified based on the darker colored neurons surrounding them. Cluster 5 includes a large cluster with documents consisting of concepts of diseases like `obesity', `diabetes', `hypertension', `hyperglycemia', and so on. This cluster also includes other diseases such as `coronary artery disease', since these diseases are highly related. The `coronary artery disease' might be an outcome of `hypertension', `hyperglycemia' or their combination. Cluster 6 contains documents discussing infections related to diseases such as `Malaria', and other respiratory infection diseases such as `tuberculosis' and `bronchitis' are also included in this cluster. 
Cluster 8 is a smaller cluster which mainly includes chest infection related documents. Cluster 7 is another smaller cluster about concepts of infections such as avian influenza. Cluster 6, 7 and 8 are close to each other since they are all about infections, but each is focused on a smaller areas of infections. Cluster 4 contains three neurons which present three different types of concepts. The left region is dominated by documents which discuss `dysphagia' and similar concepts such as `laryngospasm'. The concepts in the right half of the cluster include spinal disorders like `stenosis', `scoliosis' and `spinal instability'. The top of the cluster is dominated by pain related diseases and syndromes. Cluster 3 contains documents that are related to `paralysis' and damage of nerves. Many of them discuss paralysis of the face, hands (`carpel tunnel syndrome'), spine (`spinal cord atrophy'), brain (`celebral palsy'), legs (`spastic foot'), and so on. All these concepts are related and close to each other, and thus the cluster is well formed. Cluster 2 is a cluster of documents with concepts about different cardiovascular diseases such as `hypertension', `myocardial infarction', `coronary artery disease', `coronary heart disease', and  `ischemic strokes'. This cluster also consists of a few documents that talk about brain strokes arising from lesions in the brain, or lead to speech disorders. More than half of the documents in this cluster discuss strokes and closely related cardiac concepts. One interesting finding is that cluster 1 contains all the documents in which the only one concept is identified by UMLS MetaMap is `stroke'. However, further analysis shows that these documents have nothing to do with `stroke' as a disease. This shows the UMLS MetaMap cannot always accurately map all concepts to the semantic types. 

Figure \ref{OHSU} shows the clustering results of Ohsumed collection based on the concepts of diseases that are extracted. Based on the document frequencies of the concepts of Ohsumed collection, there are 1108 concepts out of the total 1449 that occurs only in one document and there are total 1414 concepts occur in less than five documents. That means the weights of these concepts rely heavily on the similarity measurement between the concepts.

Based on the original data set description, all documents are related to cardiovascular diseases. This lead to the shorter distances between neurons which is reflected by the color of the U-matrix. By analyzing the U-matrix and hit histogram of the trained map,  8 clusters are identified. A majority of the documents in cluster 7 are about infections and infectious diseases, with half of them from the categories of bacterial infections and mucoses (C01), virus diseases (C02) and parasitic diseases (C03). The rest of the documents from this cluster discuss other infections from categories like respiratory tract diseases (C08) and digestive system diseases (C06). There are also a few documents from immunologic diseases (C19) in this cluster. Notably, all of the documents talk about infections of different types. Cluster 1 has documents that discuss diseases about the nervous system. Whereas, the documents in cluster 2 discuss neoplasms which include different types of cancers of the brain, prostate, neck and so on. The documents in this cluster are from all the categories except virus diseases (C02) and diseases of environmental origin (C21). Cluster 3 includes documents about diseases related to hormone secretion and distribution. This cluster also includes diseases of the bones and blood, since these concepts are closely related. Cluster 4 contains documents with diseases about the ear, nose, throat, head and surrounding areas of the face. Cluster 5 is the smallest cluster and the documents concentrate on different types of tuberculosis and sexually transmitted diseases like AIDS, HPV, etc. Documents about `cryptococcosis', which is often seen in patients with HIV whose immunity has been lowered, also fall in this cluster. Cluster 6 consists of documents with concepts relating to diabetes. Documents containing concepts like `nephropathy', `impaired glucose tolerance',`non-insulin dependent diabetes' are in the left half of the cluster. Whereas, the right half of the cluster is dominated by documents with concepts such as `Crohn's disease', `renal ulceration', and `kidney stone'. Cluster 8 has documents about diseases related to different heart conditions and obstruction in the flow of blood. Since the theme of documents in Ohsumed collection is cardiovascular concepts, this cluster has documents from all of the categories except parasitic diseases (C03), neoplasms (C04) and digestive system diseases (C06).

It is worth noting that only concepts of diseases are extracted from both data sets and used for document clustering. While the original category labels of the Ohsumed collection might not be assigned based on the concepts of diseases, thus, these labels are used to evaluate the clustering performance.

Overall, the proposed document clustering and visualization framework works well on both data sets. Although Ohsumed collection has much more concepts diseases, and the majority of them have very low document frequency. On the U-matrix, the colors of the neurons surrounding the clusters demonstrate how separated these clusters are. The darker the color is, the more separated they are. That means clusters are more unrelated. The clusters on the U-matrix of the PMC-OA appear to be more separated than those of the Ohsumed collection. One reason could be that all documents are related to cardiovascular diseases, so the clusters locate more closely on the U-matrix.

\section{Conclusion and Future Work}
In this paper, a biomedical document clustering framework based on concepts of diseases is proposed. The concepts of diseases are identified by using UMLS MetaMap. Instead of using an existing ontology to generate concept representation, the concepts are represented by using vectors based on a combination of TF-IDF and Word2Vec models. The proposed similarity measure is based on the vector representations of the concepts and shows that closely associated concepts of diseases have higher similarity scores than others. A representation of documents that considers the local content and semantic similarity between the concepts within the documents is used. A weighting scheme using TF-IDF combined with similarity score between the concepts is proposed. Instead of focusing on clustering performance evaluation, clustering visualization is explored in this research. Self-Organizing Map is a clustering algorithm that provides a visualization aid to understand the clusters and distribution of the clusters, and is thus used in this research. The results show that the clustering occurs along concepts of similar nature, of similar area and organs of the body, and concepts which are synonymous to one another. Nearby clusters are related in most cases, as well. This kind of visualization will help researchers explore related articles based on concepts of diseases.

Potential future work includes visualizing clusters of larger corpora by using a hierarchical clustering architecture, evaluating this visualization aid for the task of biomedical document search and extending this framework to biomedical document clustering based on concepts of symptoms and treatments. 

\bibliographystyle{ACM-Reference-Format}
\bibliography{sigproc} 

\end{document}